\pdfoutput=1

\documentclass[11pt]{article}

\usepackage[final]{acl}

\usepackage{times}
\usepackage{latexsym}

\usepackage[T1]{fontenc}

\usepackage[utf8]{inputenc}

\usepackage{microtype}

\usepackage{inconsolata}

\usepackage{amsmath}
\usepackage{amsfonts}
\usepackage{amssymb}
\usepackage{booktabs}
\usepackage{graphicx}
\usepackage{tabularx}
\usepackage{soul}
\usepackage{comment}
\usepackage{multirow}
\usepackage[dvipsnames]{xcolor}

\title{A Study of Nationality Bias in Names and Perplexity using Off-the-Shelf Affect-related Tweet Classifiers}

\author{Valentin Barriere \\
  Universidad de Chile -- DCC, CENIA \\
  Santiago, Chile \\
  \texttt{vbarriere@dcc.uchile.cl} \\\And
  Sebastian Cifuentes \\
  CENIA \\
  Santiago, Chile \\
  \texttt{sebstian.cifuentes@cenia.cl} \\}

\begin{document}
\maketitle
\begin{abstract}

In this paper, we apply a method to quantify biases associated with named entities from various countries. We create counterfactual examples with small perturbations on target-domain data instead of relying on templates or specific datasets for bias detection. On widely used classifiers for subjectivity analysis, including sentiment, emotion, hate speech, and offensive text using Twitter data, our results demonstrate positive biases related to the language spoken in a country across all classifiers studied. Notably, the presence of certain country names in a sentence can strongly influence predictions, up to a 23\% change in hate speech detection and up to a 60\% change in the prediction of negative emotions such as anger. We hypothesize that these biases stem from the training data of pre-trained language models (PLMs) and find correlations between affect predictions and PLMs likelihood in English and unknown languages like Basque and Maori, revealing distinct patterns with exacerbate correlations. 
Further, we followed these correlations in-between counterfactual examples from a same sentence to remove the syntactical component, 
uncovering interesting results suggesting the impact of the pre-training data was more important for English-speaking-country names.  
Our anonymized code is \href{https://github.com/valbarriere/biases_ppl/}{available here}.
\end{abstract}


\section{Introduction}  

\begin{figure*}
    \centering 
    \hspace*{-.2cm} 
    \includegraphics[width=1.01\textwidth]{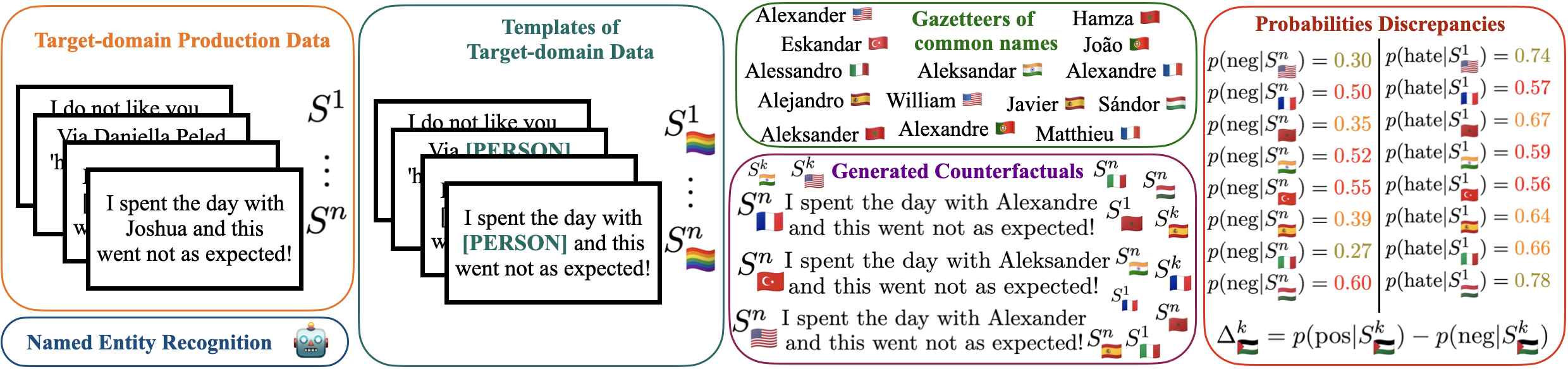}
    \caption{Overview of the counterfactual example creations. 
    We show examples with sentiment and hate speech for variation of the name "\textit{Alexander}" and two sentences $S^1$ and $S^n$. $S^1:$ "\textit{I do not like you [PER] you fucking bitch}". 
    The \textcolor{MidnightBlue}{\textbf{NER}} is applied to the \textcolor{YellowOrange}{\textbf{production data}} to create \textcolor{PineGreen}{\textbf{templates}}, which are then filled randomly with most common names from \textcolor{OliveGreen}{\textbf{gazeeters}} of different countries to create a pool of \textcolor{Purple}{\textbf{counterfactuals}}. 
    The \textcolor{BrickRed}{\textbf{discrepancies}} in probabilities is quantified using metrics such as $\Delta$. 
    }
    \label{fig:overview} \vspace*{-.1cm}
\end{figure*}

Recent trend in Natural Language Processing research, like in works published at conference such as ACL \cite{ACL23proceedings}, is to provide open-source data and models  
\cite{Scao}. This practice not only enhances its value for general research purposes but also facilitates the deployment of these models in diverse operational settings by companies or stakeholders. 
Applications such as customer experience, CV screening, Social Media analyses and moderation are example of applications that will directly impact the users in different ways. For this reason, the models applied at large scale should be scrutinized in order to understand their behavior and should tend to be fair by passing successfully a series of test to reduce their biases toward various target groups. 
Past study \citep{Ladhak2023} showed that PLMs are impacted by names, and \citet{barriere-cifuentes-2024-text-classifiers} proposed a method to quantify this to detect biases of the model toward specific countries, using the country most common names as a proxy. We are showing in this paper that this bias is systematic in several widely-used off-the-shelf classifiers on English data, and propose a method to directly link the bias level with the perplexity of the PLM
\paragraph*{Contributions}
We propose an investigation into biases related to country-specific names in widely used off-the-shelf models \citep{Camacho-collados2020,Barbieri2021}, commonly deployed in production environments for Twitter data.\footnote{Regarding the number of monthly downloads of \texttt{cardiffnlp} models from \citet{Camacho-collados2020,Barbieri2021} in the Huggingface Model Hub at the time of writing  ($>$4m for sentiment).} Our analysis reveals distinct biases in sentiment, emotion, and hate speech classifiers, showing a propensity to favor names from certain countries while markedly disfavoring those from less Westernized nations, often by a large margin. Furthermore, we establish a global-level correlation between the perplexity of associated PLMs and model predictions across both known and unknown (i.e., Out-of-Distribution; OOD) languages, demonstrated through examples in English, Basque, and Maori. At a local level, we mitigate the influence of syntax on perplexity by examining the correlation among counterfactual examples generated through minor perturbations. Notably, our findings suggest that the frequency of a name's occurrence during the training phase directly impacts the sentiment model's tendency to produce positive outputs, which highly disadvantage the non-English (i.e., OOD) persons in a world where English is widely utilized as pivot language. \textit{\textbf{Our method is unsupervised, moreover it can be applied to any classifier and any dataset}}.
\section{Related Work}

As it is known that models still learn bias when fine-tuned on downstream tasks and that the correlation is low between the intrinsic bias scores of the initial model and its extrinsic bias scores after fine-tuning \citep{Kaneko2022b,Kaneko2024}, we use a method to evaluate an already trained classifier and not the pre-trained language model. 
Some works propose such thing as general "unit-test" for NLP models \cite{Ribeiroa} or even applying a battery of fairness tests \cite{Nozza2022}. However, extrinsic methods mainly relies on template or datasets \cite{Czarnowska2021,Kurita2019a,Guo2021a}, which have been proven to influence considerably the bias estimation and conclusion across template modification \cite{Seshadri2022}. 
%
A potential solution is to apply perturbation on the test data. Perturbations can be used for attribution methods \citep{Fel2023}, but also for testing a model's robustness \citep{Ribeiroa}. They allow getting rid of the aforementioned template issue and data collection methodology: directly used on the target domain data, it prevents for not properly evaluating the intended notion of bias \citep{Blodgett2020}.

The origin of the bias generally comes from the training data \citep{Caliskan2017}, as a lot of information can be stored in the network 
\cite{Petroni2019,Carlini2021,Carlini2018} due to repetitions of the same sentences or concepts. 
%
This type of over-representation in the training data involve a representation bias, such as the one demonstrated by \citet{Kaneko2022a} regarding the gender as masculine was over-represented. 
This was found out to be correlated with the likelihood of the model. For example, \citet{Barikeri2021} propose a perplexity-based bias measure meant to quantify the amount of bias in generative language models along several bias dimensions. 
For this reason, \citet{Kaneko2022} propose to use the likelihood as a proxy to estimate the bias on gender. In our case, we validate that the bias is already present in the PLM, by calculating the correlation between the likelihood and different classes for country-name. This technique is even more efficient with generative models \citep{Ouyang2022instruct,Jiang2024} as one can apply it directly on production model. 

Although names are not inherently linked to a specific nationality, research has revealed the presence of nationality biases within them
. Delving into this underexplored domain, \citet{Venkit2023} shed light on the influence of demographic 
on biases associated with countries in language models. \citet{An2023} offer insights into the intricate relationship between demographic attributes and tokenization length, particularly focusing on biases related to first names. \citet{Zhu2023} propose 
to mitigate name bias by disentangling it from its semantic context in machine reading comprehension tasks. 
\citet{Ladhak2023} investigate the propagation of name-nationality bias, demonstrating through intrinsic evaluation with templates how names and nationalities are intrinsically linked and how biases manifest as hallucinations. 
Lastly, \citet{barriere-cifuentes-2024-text-classifiers} showed that using names as proxy works to detect country-related biases depends on the sentence's language, in multilingual sentiment and stance recognition models \citep{Barriere2023,Barriere2022,Barriere2022debating}.  
\section{Method} 

\begin{table*}
    \centering
    \hspace*{-.5cm}
    \resizebox{1.0\textwidth}{!}{
\begin{tabular}{l|cccc|cccc|cc|cc}
\multirow{2}{*}{\textbf{Country}} & \multicolumn{4}{c}{\textbf{Sentiment}} & \multicolumn{4}{c}{\textbf{Emotion}} & \multicolumn{2}{c}{\textbf{Hate}} & \multicolumn{2}{c}{\textbf{Offensive}} \\
{} & $\Delta$ & $-$ & $\approx$ & $+$ &  Joy & Opt. & Anger & Sad. & Non-hate &  Hate & Non-off. & Off. \\
\hline \hline
United Kingdom &           -1.43 &       5.4 &     1.3 &     -4.6 & -2.1 &      0.6 &   2.7 &     6.4 &     -0.2 &  23.5 &          -0.4 &       4.8 \\
United States  &           -1.35 &      5.0 &     1.7 &     -4.9 & -2.3 &     -0.5 &   4.0 &     6.5 &     -0.2 &  22.0 &          -0.5 &       6.1 \\
Canada         &           -1.43 &      5.5 &     1.5 &     -5.0 & -1.6 &     -0.2 &   2.3 &     5.0 &     -0.2 &  21.0 &          -0.4 &       4.5 \\
Australia      &           -1.37 &      5.7 &     1.2 &     -4.7 & -2.3 &      0.9 &   3.2 &     6.6 &     -0.2 &  23.0 &          -0.3 &       4.3 \\
South Africa   &           -1.58 &      5.9 &     1.2 &     -4.8 & -1.5 &      0.4 &   1.0 &     6.1 &     -0.2 &  22.5 &          -0.3 &       3.9 \\ \hline \hline
India          &           -2.70 &      7.9 &    -0.1 &     -4.4 & -2.5 &     -6.1 &   8.7 &     5.0 &     -0.1 &  10.0 &           0.1 &      -1.6 \\
Germany        &           -2.14 &      6.4 &     1.3 &     -5.3 & -0.0 &     -4.8 &  -0.2 &     4.7 &     -0.1 &  19.0 &          -0.3 &       3.3 \\
France         &           -1.58 &      7.7 &    -0.2 &     -4.0 &  0.9 &     -5.1 &  -2.5 &     3.8 &     -0.1 &  10.5 &          -0.0 &       0.1 \\
Spain          &           -2.46 &      6.0 &     2.6 &     -6.5 &  1.7 &    -13.0 &  -0.4 &     2.7 &     -0.0 &   6.0 &          -0.2 &       2.7 \\
Italy          &           -1.98 &      7.1 &     1.1 &     -5.4 &  2.5 &    -15.5 &  -0.9 &     1.5 &     -0.1 &  12.5 &          -0.2 &       2.5 \\
Portugal       &           -2.30 &      6.9 &     1.6 &     -5.9 &  1.9 &    -12.9 &   1.1 &    -0.4 &     -0.1 &   9.5 &          -0.1 &       1.8 \\
Hungary        &           -2.26 &      4.9 &     2.7 &     -6.1 &  2.4 &    -17.2 &  -1.4 &     4.0 &     -0.1 &   6.5 &           0.2 &      -2.1 \\
Poland         &           -2.02 &      3.4 &     3.6 &     -6.3 &  2.0 &    -13.7 &  -2.4 &     5.1 &     -0.1 &   9.5 &           0.1 &      -1.3 \\
Turkey         &           -2.33 &      6.8 &     0.7 &     -4.7 &  0.2 &    -11.9 &   4.8 &     1.7 &     -0.1 &   7.5 &           0.0 &      -0.3 \\
Morocco        &           -2.04 &      4.2 &     2.4 &     -5.2 & -9.0 &    -33.2 &  60.3 &   -17.4 &     -0.0 &   2.0 &           0.4 &      -4.9 \\
\bottomrule
\end{tabular}
}
\caption{Changes in probability output ($\Delta$) and in percentage of examples in each of the predicted classes, both relative to the original unmodified sentence to compare with the model's likely real-world production settings.} 
\label{tab:all_res} 
\end{table*}

We first rely on Named Entity Recognition (NER) to create counterfactual examples from the target-domain, specific of target groups, following the methodology of \citet{barriere-cifuentes-2024-text-classifiers}. The bias is assessed by quantifying the differences in the model outputs. Second, we ran a series of experiences studying the correlation between the output variations and the perplexity. 
Figure \ref{fig:overview} shows an overview of the bias detection.

\subsection{Perturbation-based Counterfactuals} 

\paragraph*{Counterfactual Generation}
A set of counterfactual examples are constructed from the target-domain data using a NER system combined with a list of most common names from different countries. Each named entity automatically tagged as person is substituted by a random common name from a specific country. 
Note that the original entity is conserved, by looking in our gazeeters its corresponding gender. 
More details are found in the Appendix \ref{app:cf_creation}. 

\paragraph*{Bias Calculation} 
In order to assess the bias, we calculate the percentage of change in terms of tagged examples, using the confusion matrices. For sentiment, we also computed the change in difference in probability between positive predictions and negative predictions $\Delta$.     

\subsection{Perplexity and Likelihood}

\paragraph*{General and Pseudo-Perplexity}
The perplexity of a language model measures the likelihood of data sequences and represents how fluent it is \cite{Carlini2018}. In simpler terms, perplexity reflects how unexpected a particular sequence is to the model. A higher perplexity suggests that the model finds the sequence more surprising, while a lower perplexity indicates that the sequence is more likely to occur. 
We refer to the definition of pseudo-log-likelihood introduced by \citet{Salazar2020}, the pseudo-perplexity being the opposite of it. 
For a sentence $S = w_1, w_2, . . . , w_{|S|}$, the pseudo-log-likelihood ($PLL$) score given by Eq. \ref{eq:PLL}, can be used for evaluating the preference expressed by an MLM for the sentence $S$.

\begin{equation}
    PLL(S)= - \sum_{i=1}^{|S|} \text{log} P_{MLM}(w_i|S_{\setminus wi} ; \theta)
    \label{eq:PLL}
\end{equation}

The Log-Perplexity as defined in \citet{Carlini2018} is the negative log likelihood, hence \textbf{we use pseudo-log-perplexity as simply the opposite of the PLL}.\footnote{Contrary to the definition of \citet{Salazar2020} defining it on a complete corpus, summing between all the sentences before passing it to exponential.} 
More details are provided in Appendix \ref{app:likelihood}. 
In the following, we will not use the term pseudo- when talking about the pseudo- perplexity or likelihood. 

\paragraph*{Bias quantification}
We calculated the Pearson correlation between the probabilities output and likelihood in two ways. First, what we call \textit{global} correlation, i.e., between all the examples of the dataset, in order to shed lights on a general pattern between perplexity and subjectivity. 
Second, what we call \textit{local} correlations, i.e., between elements coming from the same original sentence, before averaging them. 
In this way, we can disentangle the syntactic aspect of the sentences that have an impact in the likelihood calculation. This is similar to normalizing the perplexity and likelihood of every examples coming from the same sentence before calculating the Pearson correlation.  

\section{Experiments and Results} 

\subsection{Experiments} 

\paragraph{Bias Detection}
Our first experiment focuses on quantifying the country names bias for different off-the-shelf models previously learned on tasks that are related to affects, looking at the probability of positiveness and the percentage of change in number of predicted examples per class. 

\paragraph{Global Perplexity}
The second experiment aims to show that the model predictions are in general intricately linked with the perplexity even for unknown languages. 
We first create datasets in these unknown languages using Machine Translation (MT) in order to preserve the semantic content in-between the different languages, as they did in in \citet{Balahur2013}. 
We then calculate the "global" correlation between perplexity and output probabilities in English and unknown languages such as Maori and Basque, which we obtain using Google Translate.\footnote{Google MT is based on the LLM PaLM 2 \citep{Google2023}, which should work reasonably well for these two languages already used in production.}
More details in Appendix \ref{app:MT}. 

\paragraph{Local Perplexity}
To remove the syntactic aspect influencing both perplexity and predictions, we conduct experiments focusing on what we call "local" correlation, which is between the relative probabilities of each class among counterfactual examples (i.e., generated with minor perturbations) and their associated relative perplexity. 

\subsection{Experimental Protocol}

\paragraph{Gazeeters}
We used the dataset collected from Wikidata Query Service.\footnote{\url{https://query.wikidata.org/}} by the authors of \texttt{Checklist}, composed of common first and last names as well as the associated cities from several countries. This makes a total of 16,771 male first names, 12,737 female first names, 14,797 last names 
from 194 countries. 

\paragraph{NER}
We use a multilingual off-the-shelf NER system available on the Spacy library \cite{spacy} and created for social media (named \texttt{xx\_ent\_wiki\_sm}) to identify entities for removal in target-domain data, aligning with the data used during model deployment. 

\paragraph{Perturbation}
For every sentence $x$, we create $50$ random perturbations of this sentence for each of the target countries. 

\paragraph{Dataset}
In order to apply our method to data similar to production data, we collected 8,891 \textbf{\textit{random tweets in English}} by using the IDs from the Eurotweets dataset \citep{Mozetic2016}. 
%
The 8,891 tweets used in the experiment correspond to a random selection of 10\% of the English tweets of the EuroTweets dataset \citep{Mozetic2016} downloaded in June 2020.\footnote{No label were used.} 

\paragraph{Tested Classifiers}
The models used were the ones of \cite{Camacho-collados2020,Barbieri2021} for multilingual sentiment analysis, monolingual hate speech, emotion recognition and offensive text detection:
\small{\texttt{cardiffnlp/twitter-xlm-roberta-base-sentiment}, \texttt{cardiffnlp/twitter-roberta-base-hate}, \texttt{cardiffnlp/twitter-roberta-base-emotion},} \normalsize{and }\small{\texttt{cardiffnlp/twitter-roberta-base-offensive}}. \normalsize 
%
Experiments were run using Tensorflow 2.4.1 \cite{Abadi2016}, transformers 3.5.1 \cite{Wolf2019}, a GPU Nvidia RTX-8000 and CUDA 12.0. 
\subsection{Results} 

\paragraph*{Bias Detection}
Table \ref{tab:all_res} provides a comprehensive overview of the impact of country-specific named entities on sentiment, emotion, hate speech, and offensive text classifications across diverse classifiers. Notably, it reveals significant variations in model predictions based on the presence of different country names within textual data. 
For sentiment analysis, it is striking to observe substantial shifts in sentiment probabilities ($\Delta$)\footnote{$\Delta$'s standard deviations are proportional to its values.} across countries. For instance, countries like India, Turkey or Spain 
exhibit noteworthy deviations in sentiment probabilities, indicating potential biases in classifier outputs concerning specific national contexts.\footnote{This is interesting as Spanish (resp. Indian dialects) are the main foreign languages of migrants in US (resp. UK).} The percentages of predicted negative, neutral, and positive sentiments further underscore the nuanced nature of these biases, with certain countries consistently receiving more positive or negative sentiment classifications compared to others. 
Emotion analysis reveals intriguing patterns in the distribution of predicted emotions across countries. Optimism shows an interesting pattern where the non-English names highly decrease this prediction, up to -33\% for Moroccan. It is also notable that Moroccan names provoke a very high increase (60\%) of anger predictions at the expense of the other classes. 
%
%
Finally, a similar pattern can be seen for the hate speech and offensive text classifiers. English-speaking countries names highly favor hate speech detection, even as a false positive, compared to other countries. For offensive text detection, there is an increase of 6.1\% with counterfactuals using US names and a decrease of 4.9\% and 2.1\% using Moroccan and Hungarian names. 


\begin{table}
    \centering
    \hspace*{-.2cm}
    \resizebox{.49\textwidth}{!}{
\begin{tabular}{lc|ccc}
\textbf{Task} & \textbf{Label}  &\textbf{English} &        \textbf{Basque} &        \textbf{Maori}\\ \hline \hline
\multicolumn{2}{l|}{Hate} &  3.17 &  23.07 &  22.31 \\ \hline
\multirow{3}{*}{Sentiment} & $-$ & -11.39 &  25.48 &  35.33 \\
    & $\approx$ &  19.27 &  -19.98 &  -36.23 \\
      & $+$ & -5.41 & -3.04 &  5.86 \\
\bottomrule
\end{tabular}
}
\caption{Global correlations between PPL and classes for different languages, tasks or pre-trainings.} 
\label{tab:PPL_gen}
\end{table}

\paragraph*{Global Subjectivity-Perplexity Correlation}
Table \ref{tab:PPL_gen} shows the correlations between the perplexity and the labels for Sentiment and Hate speech tasks using tweets from different languages, obtained using Machine Translation. 
%
For the hate speech model, the global correlation between the hate speech class and the perplexity is almost close to zero for English data, which is good since showing no spurious pattern between perplexity and hate speech prediction. 
However, the correlations are higher for the unknown language such as Basque and Maori, where it reaches more than 22\%. The model tends to classify as hate speech more easily texts having a higher perplexities, i.e., that are outside the training distribution. 
%
For the Sentiment model, the pattern for Basque and Maori language is the same, high positive/negative correlation for the negative/positive class, which means that the less the sentence is similar to the train distribution, the more negative it would be. Additional experiments using other 
languages are confirming the results, and are available in Appendix \ref{app:exp2}.

\begin{table}
    \centering
    \resizebox{.4\textwidth}{!}{
    \begin{tabular}{l|ccc}
     \multirow{2}{*}{\textbf{Country}} & \multicolumn{3}{|c}{\textbf{Sentiment}} \\
     & $-$ & $\approx$ & $+$ \\
    \midrule
    United Kingdom &    15.03 &    5.89 &   -18.26 \\
    United States  &    14.70 &    6.63 &   -18.41 \\
    Canada         &    15.18 &    4.91 &   -17.68 \\
    Australia      &    15.68 &    5.46 &   -18.52 \\
    South Africa   &    13.12 &    5.87 &   -16.67 \\
    India          &     7.64 &    5.18 &   -11.75 \\
    Germany        &    13.62 &    4.50 &   -16.34 \\
    France         &     8.18 &    4.42 &   -11.47 \\
    Spain          &    11.37 &    4.16 &   -14.23 \\
    Italy          &    11.09 &    3.79 &   -13.57 \\
    Portugal       &     9.45 &    2.93 &   -11.97 \\
    Hungary        &     8.37 &    2.89 &   -10.79 \\
    Poland         &     9.88 &    3.22 &   -12.32 \\
    Turkey         &     9.62 &    2.79 &   -11.86 \\
    Morocco        &     9.07 &   -0.16 &    -8.25 \\ \hline \hline
    \textbf{Overall}            &    11.17 &    4.63 &   -14.40 \\
    \bottomrule
    \end{tabular}
    }
    \caption{Correlations between the relative perplexity of the model and the relative output probabilities.
    } 
    \label{tab:PPL_local}
\end{table}
\paragraph*{Local Subjectivity-Perplexity Correlation}
Table \ref{tab:PPL_local} shows correlations between the relative perplexity of the model and the probabilities of different classes. The results are very different from global correlations. 
Notably, there is a negative correlation between perplexity and positiveness of the sentiment, which implies that names that are more similar to what was seen during the PLM pre-training will imply a more positive output of the sentiment classifier. This trend is particularly pronounced among English-speaking countries. 
%
Due to lack of space, more details and results can be found in Appendix \ref{app:exp3}. 

\section{Conclusion} 

Bias at the nationality level can also occur with the most common entities of the country such as names. We show its occurrence in this paper for a set of tasks that are related with affect and subjectivity classification, using several transformer models widely used on Twitter data
 . Motivated by prior research, we studied the link between this bias and the perplexity of the PLM showing \textit{(i)} exacerbate correlations in unknown languages, and \textit{(ii)} verify that correlation can be related to names using counterfactual sentences. 
We found out 
interesting patterns using the Pearson correlations between the classes and perplexity, revealing higher correlations for English-speaking country names, meaning that the exposition bias on names impacts the predictions also in-between a country. 
%

\section{Limitations}

First, our method only relies on Named Entities, so it does miss all the implicit hate speech. Nevertheless, it is a system with low recall but high precision as when it detects a change, meaning that the classifier behavior is biased. 
Second, even if our method slightly perturbates the data from the target distribution, it does not explicitly keep it inside, creating examples that might be a bit outside the distribution of the production data. We think that is the reason why we see a general shift toward a more negative sentiment when comparing perturbated examples and true examples (negative predictions always augment while positive predictions always decrease). It would be more natural to use target-data-specific lexicons, or use a generative model to do the job. However, we think that this is a fair comparison toward all the countries and it can drive a pertinent conclusion on a relative bias between the different countries. 
Another bias induction can also come from the fact that some names can be non gendered in some context, such as Claude as a first-name or Jane as a surname (for a man) that would be tagged as feminine. Co-reference resolution could mitigate this issue, even though we believe it is uncommon. 
Finally, we compare a masked language model, but further experiments are left for future work
using generative models such as flan-T5 \citep{Chung2022} or Mixtral \citep{Jiang2024} where the same model computes both label and perplexity, for example using label tokens probabilities to estimate the probabilities \citep{Hegselmann2023}.  


\section*{Acknowledgements}
The authors thank the reviewers for the various comments that helped to improve the manuscript. 
This work has been partially funded by National Center for Artificial Intelligence CENIA FB210017, Basal ANID.

\bibliography{JRC.bib}

\appendix

\begin{table*}[h!]
\centering
\begin{tabular}{c|cccccc|cc}
 \textbf{Label} & \textbf{English} & \textbf{Dutch} & \textbf{Spanish} & \textbf{Hindi} & \textbf{Malayalam} & \textbf{Turkish} & \textbf{Basque} & \textbf{Maori} \\ \hline  \hline
$-$           & -11.39    & -13.87         & -6.28            & -10.89         & -7.03              & -6.02       &  25.48 &  35.33      \\ 
$\approx$      &  19.27       & 21.61          & 19.00          & 25.54            & 9.12           & 16.54       &  -19.98 &  -36.23       \\ 
$+$          & -5.41     & -7.13          & -11.10         & -13.50           & -1.94          & -10.32       & -3.04 &  5.86      \\ \bottomrule
\end{tabular}
\caption{Global correlations between PPL and classes
for different languages using the multilingual sentiment model}
\label{tab:app_PPL_gen}
\end{table*}

\section{Counterfactual Examples Creation} \label{app:cf_creation}

\paragraph{Notation}
We decide to slightly change the notations of \citet{Czarnowska2021} because our target groups are country-related, which can be defined by different attributes such as names of persons or locations.   
We use $\mathcal{A}$ as a set of target words sets such that $\mathcal{A} = \{A_1, A_2, ..., A_{|T|}\}$ where $A_t$ represents the target words set of the target group $t$ for the attribute $A$,\footnote{It can be name regarding the gender, surname, location,...} and $|T|$ the number of target groups that we consider. The set of source examples $\mathcal{S} = \{S^1, S^2, ..., S^{|\mathcal{S}|}\}$ contains the sentences from our target-domain data with at least one named entity (such as a person or a location), and $\mathcal{S}' = \{S'_1, ..., S'_{|\mathcal{S}|}\}$ the set of sets of perturbated examples, $S'_i=\{S^i_{t,j}, j=1..E\}$ the set of perturbated examples of the sentence i for the target group $t$, with $E$ the number of counterfactual examples. 
We use $\Phi$ as the score functions, and $d$ as the distance metrics used on top of the score functions.  

In the example in Figure \ref{fig:overview}, for simplicity reasons we show only one example of name per country, which means $j=1$ in $S^i_{t,j}$ and $t$ is represented as the flag of the country.

\begin{table*}
    \centering
    \resizebox{.9\textwidth}{!}{
    \begin{tabular}{l|ccc|cccc|cc}
    \toprule
     \multirow{2}{*}{\textbf{Country}} & \multicolumn{3}{|c|}{\textbf{Sentiment}} & \multicolumn{4}{|c|}{\textbf{Emotion}} &  \multirow{2}{*}{\textbf{Hate}} & \multirow{2}{*}{\textbf{Offensive}} \\
     & $-$ & $\approx$ & $+$ &   Anger &    Joy & Opt. & Sadness &   &  \\
    \midrule
    United Kingdom &    15.03 &    5.89 &   -18.26 &    2.02 &   6.82 &   -16.46 &   14.87 &  3.96 &      2.75 \\
    Ireland        &    11.69 &    5.78 &   -15.72 &    0.21 &   8.77 &   -15.30 &   11.78 &  2.67 &      5.20 \\
    United States  &    14.70 &    6.63 &   -18.41 &    1.99 &   8.23 &   -19.01 &   17.09 &  4.44 &      4.90 \\
    Canada         &    15.18 &    4.91 &   -17.68 &    1.62 &   7.10 &   -16.73 &   15.22 &  2.97 &      4.31 \\
    Australia      &    15.68 &    5.46 &   -18.52 &    2.06 &   7.70 &   -17.55 &   15.50 &  4.10 &      3.03 \\
    New Zealand    &    15.17 &    4.80 &   -17.65 &    3.29 &   5.95 &   -17.53 &   16.48 &  3.23 &      2.21 \\
    South Africa   &    13.12 &    5.87 &   -16.67 &    1.47 &   6.79 &   -16.26 &   14.97 &  3.67 &      3.50 \\
    India          &     7.64 &    5.18 &   -11.75 &   -0.37 & -12.23 &    10.32 &    1.84 &  2.50 &     12.03 \\
    Germany        &    13.62 &    4.50 &   -16.34 &    2.66 &   4.37 &   -12.99 &   11.61 &  2.12 &      4.15 \\
    France         &     8.18 &    4.42 &   -11.47 &    1.66 &   5.37 &   -10.79 &    7.51 &  2.59 &     10.19 \\
    Spain          &    11.37 &    4.16 &   -14.23 &    1.97 &   4.47 &    -9.59 &    6.10 & -1.16 &      2.36 \\
    Italy          &    11.09 &    3.79 &   -13.57 &    0.39 &   1.69 &    -5.67 &    6.14 & -1.92 &      0.76 \\
    Portugal       &     9.45 &    2.93 &   -11.97 &    0.51 &   3.29 &    -7.23 &    6.09 & -1.15 &      2.73 \\
    Hungary        &     8.37 &    2.89 &   -10.79 &    2.02 &  -0.57 &    -5.71 &    7.08 & -3.95 &      0.73 \\
    Poland         &     9.88 &    3.22 &   -12.32 &   -0.99 &   5.47 &    -6.72 &    3.67 & -4.45 &      6.66 \\
    Turkey         &     9.62 &    2.79 &   -11.86 &    1.25 &  -1.25 &    -5.50 &    9.02 & -2.74 &      0.73 \\
    Morocco        &     9.07 &   -0.16 &    -8.25 &    2.07 & -25.60 &    21.88 &    8.76 &  1.53 &     -4.44 \\ \hline \hline
    \textbf{Overall}            &    11.17 &    4.63 &   -14.40 &    2.77 &  -3.66 &    -5.05 &   10.61 &  1.69 &      2.38 \\
    \bottomrule
    \end{tabular}
    }
    \caption{Correlations between the relative perplexity of the model and the relative probabilities of the different classes. We only use hate and offensive speech detection as it is binary classification.} 
    \label{app:tab:PPL_local}
\end{table*}

\paragraph{Country-Specific Entities Gazeeters}

Our method is relying on country-specific gazeeters, that can be for different type of named entities: one gazeeter of a specific attribute $A$ from a given country $t$ will contain words related to this country. For example, if the name is the attribute and the country is France, we will obtain the set of the most common French names for man or woman $\mathcal{N}_{\text{France}}=\{\text{Matthieu}, \text{Jean}, \text{Sophie}, ...\}$ or last names $\mathcal{L}_{\text{France}}=\{\text{Lepennec}, \text{Fourniol}, \text{Denis}, ...\}$. 
The proposed method relies on gazeeters that are country-specific, that can be for different type of named entities. 


\paragraph{Data Perturbation}

The detected entities, in combination with attributes $\mathcal{A}$, form a dataset for generating contrastive examples $\mathcal{S}' = \{S'_1, ..., S'_{|\mathcal{S}|}\}$ related to specific target groups. The random subtraction process follows \citet{Ribeiroa} method using simple patterns and the Spacy library \cite{spacy}. 
Even though the model utilized is robust and widely employed in the industry, given the noisy nature of tweets, it may occasionally miss a name but is more likely to rightfully detect one (with lower recall but higher precision on noisy data). 
We manually examined 100 examples where a Person (PER) entity was detected in our downloaded data, and found a satisfying precision of the NER to be 88\%. Subsequently, our method utilizes as templates examples with detected names (which are pertinent templates if precision is high).


\section{Pseudo-Likelihood} \label{app:likelihood}

It noteworthy that it is possible to use other metrics such as the All Unmasked Likelihood (AUL) or AUL with Attention weights of \citet{Kaneko2022a}. 
Nevertheless, in our case we use examples from the target domain, hence we do want to take into account the bias introduced by the other unmasked token words in the context. Indeed, the models studied in this work are likely be deployed on data following the same distribution.  

\section{Machine Translation} \label{app:MT}
Google Translate was employed as MT, known for its up-to-date machine translation capabilities, although originally intended for general text rather than tweets. However, we do not see this as crucial. We did not check if the label is conserved because it is not the purpose as our method does not even use the original labels: the method in the 2nd experiments measures the correlation between output labels and tweet perplexity, whether it is in English, Maori or Basque. Our aim in utilizing MT was to maintain tweet content while creating our tweets in low-resource languages, as \citet{Balahur2013} did.

%

\section{Global Subjectivity-Perplexity Correlation} \label{app:exp2}

We extend the experiments of Table \ref{tab:PPL_gen}, using the exact same setting, but with other languges: Dutch, Spanish, Hindi, Malayalam and Turkish. We show the results in Table \ref{tab:app_PPL_gen}. It is possible to see that the sentiment model is behaving for these "known languages" the same way it behaves with English, with a negative correlations on the negative and positive sentiment and a positive correlation with the neutral sentiment. The behavior that we see for out-of-distribution languages such as Maori or Basque is very different.

\section{Local Subjectivity-Perplexity Correlation} \label{app:exp3}

Table \ref{app:tab:PPL_local} show the local correlations between the perplexity and probability outputs for all the classifiers. 
Regarding emotions, optimism and sadness show the same patterns than positive and negative sentiments. Surprising reverse trends are observed for Indian and Moroccan names in the positive emotion, which means the more (resp. less) stereotype is the name, the more it tend to classify joy (resp. optimism). 
Regarding hate speech and offensive text, the correlation are low. 
However, for hate speech we can notice that the trend is almost reverse between English-speaking and non-English-speaking countries.

\end{document}